\documentclass[letterpaper, 10 pt, conference]{ieeeconf}  

\IEEEoverridecommandlockouts                              

\usepackage{graphicx} 
\usepackage{capt-of}
\usepackage{amsmath} 
\usepackage{amssymb}  
\usepackage{bbm}
\usepackage{gensymb}
\usepackage{multirow}
\usepackage{pifont}
\newcommand{\cmark}{\ding{51}}%
\newcommand{\xmark}{\ding{55}}%
\usepackage{xcolor}

\newcommand{\te}[2]{$\text{#1}_{\uparrow \text{#2}\%}$} 

\usepackage{etoolbox}
\usepackage[singlelinecheck=false]{caption}
\makeatletter
\patchcmd{\@makecaption}
  {\scshape}
  {}
  {}
  {}
\makeatletter
\patchcmd{\@makecaption}
  {\\}
  {:\ }
  {}
  {}
\makeatother

\title{\LARGE \bf
Learning Better Representations for Crowded Pedestrians in Offboard LiDAR-camera 3D Tracking-by-detection
}

\author{Shichao Li, Peiliang Li, Qing Lian, Peng Yun, and Xiaozhi Chen
\thanks{The authors are with the Department of Perception, Zhuoyu Technology, Shenzhen, China. Email: {\tt\small nicholas.li@connect.ust.hk}}%
}
\begin{document}

\maketitle
\thispagestyle{empty}
\pagestyle{empty}

\begin{abstract}
Perceiving pedestrians in highly crowded urban environments is a difficult long-tail problem for learning-based autonomous perception. Speeding up 3D ground truth generation for such challenging scenes is performance-critical yet very challenging. The difficulties include the sparsity of the captured pedestrian point cloud and a lack of suitable benchmarks for a specific system design study. To tackle the challenges, we first collect a new multi-view LiDAR-camera 3D multiple-object-tracking benchmark of highly crowded pedestrians for in-depth analysis. We then build an offboard auto-labeling system that reconstructs pedestrian trajectories from LiDAR point cloud and multi-view images. To improve the generalization power for crowded scenes and the performance for small objects, we propose to learn high-resolution representations that are density-aware and relationship-aware. Extensive experiments validate that our approach significantly improves the 3D pedestrian tracking performance towards higher auto-labeling efficiency. The code will be publicly available at this HTTP URL\footnote{https://github.com/Nicholasli1995/PCP-MV}.
\end{abstract}

\section{INTRODUCTION}
\label{sec:intro}
The need to navigate like humans in urban environments becomes increasingly urgent for autonomous driving systems that aim for high levels of autonomy. One key challenge is how to accurately perceive crowded pedestrians that occlude each other when driving in cities. Recent state-of-the-art (SOTA) visual object perception approaches adopt a data-driven paradigm and achieve significant progress by learning deep representations in an end-to-end framework~\cite{carion2020end, misra2021end, zeng2022motr, hu2023planning, zhang2023monodetr, ding20233dmotformer}. However, these deep models are data-hungry and require a large amount of ground truth (GT) 3D trajectory annotations that are geometrically accurate and temporally coherent. These annotations are expensive as annotators need to manually label 3D objects in the captured point cloud (PC) sequences. How to speed up the labeling process with an automatic offboard reconstruction system is recently recognized as an essential problem in tackling the data bottleneck~\cite{yang2021auto4d, fan2023once, ma2023detzero}. These offboard systems perform 3D multiple-object-tracking (MOT) from LiDAR sequences and images in an offline manner and can exploit more computational power than online systems. Their inference results are supplied to annotators to reduce unnecessary manual efforts.

An offboard reconstruction system must generate accurate 3D MOT results to reduce labeling costs. Previous studies tackled the LiDAR 3D MOT problem and achieved significant progress in public benchmarks~\cite{sun2020scalability, caesar2020nuscenes}, yet did not dive deep into the scenario with highly crowded pedestrians. We found this scenario incurs a majority of labeling costs due to a large number of objects and requires a systematic study for efficiency improvement. However, several difficulties occur when addressing this problem. Due to the increased pedestrian density, occlusions frequently occur, and the captured PC may not be informative enough to distinguish adjacent pedestrians. In addition, models trained with common scenes face generalization issues when applied to crowded scenes. These problems cause enormous detection and tracking issues for an auto-labeling system and are even more severe considering the noisy and sparse PC captured from low-cost LiDAR sensors. 

\begin{figure}[t]
    \small
    \centering
    \begin{minipage}{.23\textwidth}
        \centering
        \includegraphics[width=\linewidth]{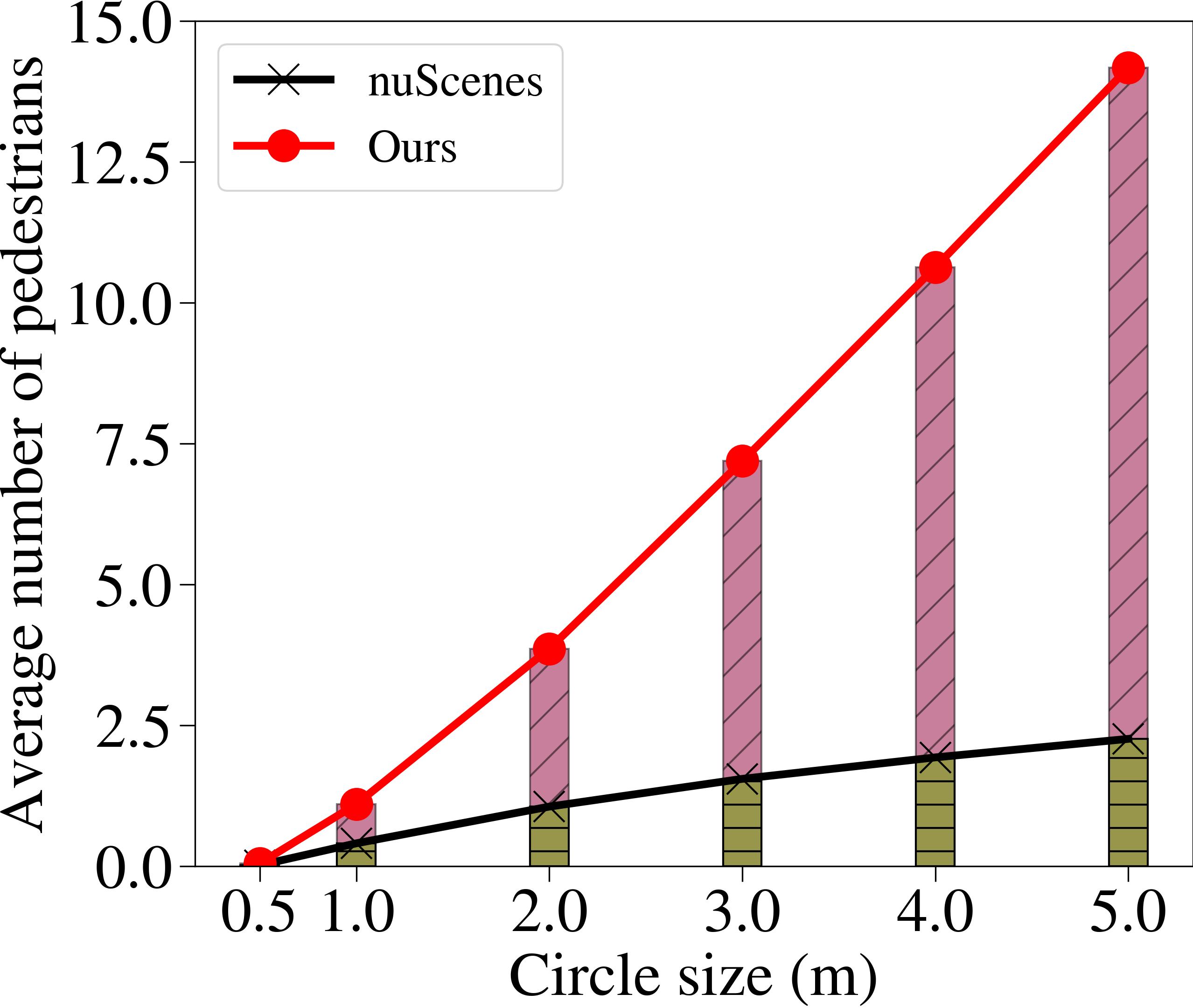}
        \captionof{figure}{A comparison of the pedestrian density with varying circle radii.}
        \label{fig:dataset_stats}
    \end{minipage}%
    \hfill
    \begin{minipage}{.24\textwidth}
        \centering
        \resizebox{0.99\textwidth}{!}{
            \begin{tabular}{lc}
            MV datasets     & Density-2 \\ \hline \hline
            nuScenes~\cite{caesar2020nuscenes}    & 0.7        \\ \hline
            H3D~\cite{patil2019h3d}       & 1.5        \\ \hline
            Waymo Open~\cite{sun2020scalability} & 1.0        \\ \hline
            PCP-MV (Ours)       & 3.8            \\
            \hline
            \end{tabular}
        }
        \captionof{table}{Comparison of the pedestrian density using a circle radius of 2 meters. The statistics of other datasets are taken from~\cite{cong2022stcrowd}.}  
        \label{tab:density_comp}         
    \end{minipage}
\end{figure}

 \begin{figure*}[h]
\centering
\includegraphics[width=1.\textwidth]{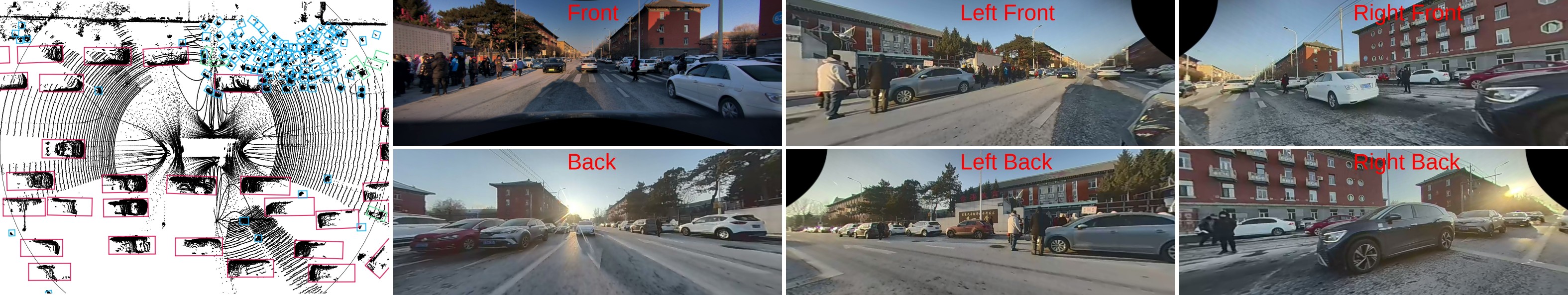}
\caption{An example frame of PCP-MV. Left: Bird's Eye View of the captured point cloud and box annotations. Right: six input camera views. Pedestrian annotations are shown in blue, and one can notice the high bounding box density. The data collection vehicle has four cameras, including two surrounding-view (SV) fisheye cameras. The fisheye images are rectified to produce four SV input pinhole images to our system. Best viewed in color and zoom in for more details.}
\label{fig:data}
\end{figure*}

An in-depth treatment of our problem requires a multi-view (MV) LiDAR-camera 3D MOT benchmark for highly crowded pedestrians in urban autonomous driving. Few existing benchmarks fit our problem setting. We thus collect a new benchmark, PCP-MV, which stands for \textbf{p}erceiving \textbf{c}rowded \textbf{p}edestrians with LiDAR and \textbf{m}ulti-\textbf{v}iew (MV) cameras. Fig.~\ref{fig:dataset_stats} compares the pedestrian density between the nuScenes~\cite{caesar2020nuscenes} validation set and PCP-MV. We compute the density by counting the number of other pedestrians within a circle of a specified radius centered on each pedestrian and average over all pedestrians. The density of PCP-MV is much higher than that of the nuScenes validation set with various radii, and Tab.~\ref{tab:density_comp} shows that PCP-MV has higher pedestrian density than other MV datasets using a radius of 2 meters. These facts make PCP-MV a more suitable benchmark for studying our problem.

This study starts with a baseline offboard LiDAR-camera 3D MOT system following a tracking-by-detection paradigm. We adopt a SOTA multi-modal 3D object detector, BEVFusion~\cite{liu2023bevfusion} for center-based tracking~\cite{yin2021center}. However, this system performs poorly in PCP-MV. For further performance enhancement, we take a representation learning viewpoint and propose a new high-resolution Bird's Eye View (BEV) representation learning approach that is aware of pedestrian density and inter-pedestrian relationships. Extensive ablation studies validate our proposed approaches, which significantly improve the 3D tracking performance of crowded pedestrians. In summary, our contributions are

\begin{itemize}
    \item We are the first to delve into the challenging offboard 3D MOT problem for highly crowded pedestrians in autonomous driving with LiDAR and MV cameras. We identify the key challenges in designing such an offboard system. 
    \item We collect a suitable benchmark for our problem and propose a new representation learning approach to address the aforementioned challenges.
    \item Our approach features high-resolution sparse representation learning along with a new density-aware loss and a new relationship-aware target. Extensive experiments validate that they significantly boost offboard 3D MOT performance for crowded pedestrians.
\end{itemize}

\section{Related works}
\noindent\textbf{LiDAR-camera 3D multiple-object-tracking.}
Estimating the trajectories of dynamic objects from LiDAR-only inputs has been studied extensively~\cite{weng20203d, weng2020gnn3dmot, yin2021center, luo2021exploring, stearns2022spot, fan2023once}. Many studies adopt a tracking-by-detection framework, where a deep-learning-based model first detects objects in each frame, and a tracker links them into trajectories. Earlier studies such as AB3DMOT~\cite{weng20203d} use motion models such as constant-velocity Kalman filters for temporal association, while later studies~\cite{zhou2020tracking, yin2021center} learn the associations with multi-frame features. Despite the progress, LiDAR-only approaches face difficulties in discriminating objects that have a small number of points caused by the sparse nature of PCs. They also fail for auto-labeling applications that require semantic object attribute estimation due to a loss of color information. A multi-modal system with MV camera sensors is thus more advantageous for general offboard auto-labeling. However, recent LiDAR-camera 3D MOT studies~\cite{chiu2021probabilistic, liu2023bevfusion} are in a more general context. Few studies have tackled the problem of challenging crowded scenarios, which is critical for high-level autonomous perception in cities, and this study aims to address this gap.

\noindent\textbf{Detecting and tracking crowded pedestrians.}
Detecting and tracking humans in crowded scenes are studied in both the computer vision~\cite{leibe2005pedestrian, zhang2016far, liu2016multi} and the robotics communities~\cite{linder2016multi}. However, most studies are for visual 2D object detection~\cite{luo2020whether, chu2020detection, wu2020temporal, lin2020graininess, li2022occlusion} and tracking~\cite{sundararaman2021tracking}. To improve 2D object detectors in a crowded environment, previous studies proposed using visual attention~\cite{pang2019mask} and visibility~\cite{zhang2018occlusion} to deal with occlusion and using improved post-processing strategies~\cite{huang2020nms, lin2020detr}. However, few studies the problem in 3D environments with LiDAR and MV cameras. Linder et al.~\cite{linder2016multi} investigated this problem before the deep representation learning era and conducted experiments in indoor environments for an online application. Our study instead aims at realistic autonomous driving in complex outdoor environments and focuses on offboard learning-based auto-labeling applications.

\noindent\textbf{Benchmarks for crowded pedestrian perception.}
There are 2D object detection~\cite{shao2018crowdhuman}, pose estimation~\cite{li2019crowdpose}, and tracking~\cite{dendorfer2020mot20} benchmarks for crowded humans, yet they do not contain 3D annotations and multi-sensor data. Previous MV autonomous driving benchmarks~\cite{patil2019h3d, liu2023bevfusion} such as nuScenes~\cite{liu2023bevfusion} provide 3D trajectory annotations, yet has a low pedestrian density as mentioned in Sec.~\ref{sec:intro}. \cite{cong2022stcrowd} proposed a 3D benchmark for crowded humans yet only has a monocular camera. Instead, we collect a new dedicated urban driving benchmark for multi-modal MV 3D MOT of very crowded pedestrians.

\begin{figure*}[t]
\centering
\includegraphics[width=1.\textwidth]{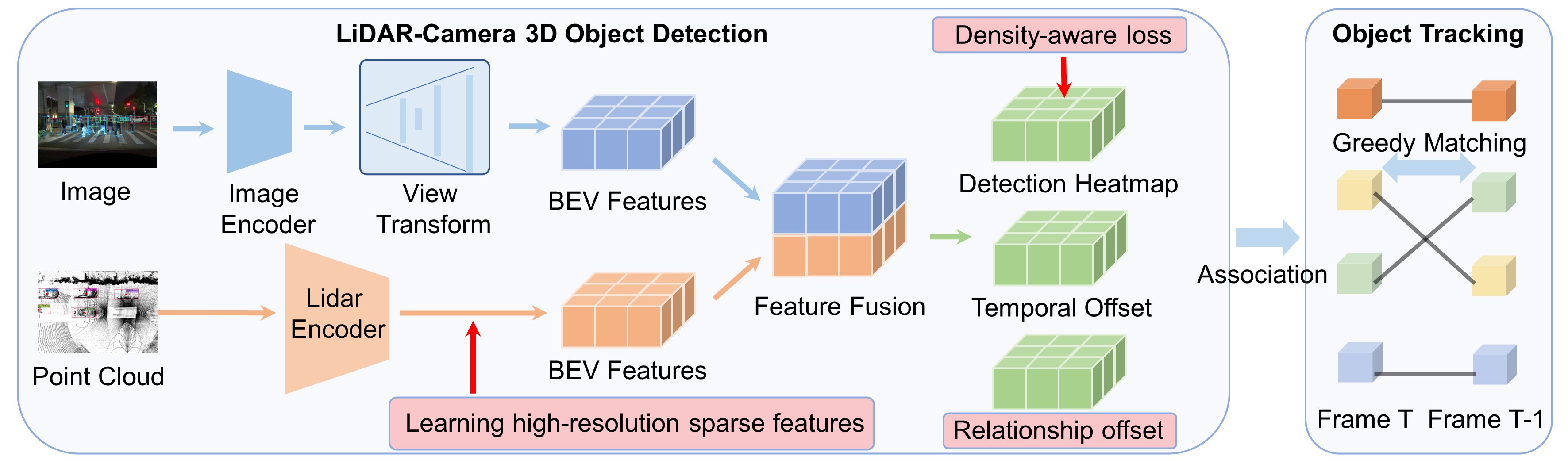}
\caption{Diagram of our offboard system for 3D MOT of crowded pedestrians. Our proposed representation learning approaches are highlighted with pink rectangles. A Swin Transformer~\cite{liu2021swin} extracts MV camera features. A uniform voxel grid is built in 3D space, where each voxel gathers features at its projected location on the images. The visual voxel features are pooled in the height dimension to form a BEV feature. The point cloud inputs are voxelized and processed by sparse convolution layers. A CNN fuses LiDAR and camera BEV features, and a head module predicts object attributes and offsets used for tracking. To enhance representation learning for crowded pedestrians, we further propose the density-aware loss, the relationship offset targets, and the high-resolution sparse feature learning module.}
\label{fig:framework}
\end{figure*}

\section{Data collection and labeling}
Due to the shortage of MV LiDAR-camera dataset for highly crowded pedestrians, we collect new driving data ourselves. Our data collection vehicles are equipped with livox LiDAR sensors and MV cameras that cover 360$^{\circ}$ field of view. Our data collectors drive in urban areas in China with heavy traffic and manually select data sequences containing a high density of pedestrians. Our annotators first exhaustively label objects that have at least one point. They then verify the correctness and temporal identity coherence of the 3D annotations in both point cloud and image views. Fig.~\ref{fig:data} shows one frame of our collected data and annotations. We split the annotated data into a training split and a test split, where the data in the two splits are recorded from different sequences. The training split contains 123 sequences with 9,552 frames. The test split is our PCP-MV benchmark with 1,414 frames from 18 sequences.

\section{Methods}
\subsection{System overview}
\label{subsec:overview}
Our offboard system is deployed via cloud-based GPU clusters to pre-label requested data before manual modification, speeding up the labeling efficiency. It takes LiDAR sequences, MV RGB images, ego poses, and calibration parameters as inputs, and reconstructs the object trajectories. Specifically, it processes $N$ frames of point cloud and images $\left\{ (\mathcal{P}_t, \mathcal{I}_t^i)  \ | \ t=1,2,...,N, i=1,2,...,M \right\}$. The data in each frame consists of $n_t$ points as $\mathcal{P}_t \in \mathbb{R}^{n_t \times 4}$ encoding point location and intensity and $M$ RGB images as $\mathcal{I}_t^i \in \mathbb{R}^{H \times W \times 3}$. The system performs tracking-by-detection where a learning-based detector detects objects for each frame as $m_t$ bounding boxes $\hat{b}_t \in \mathbb{R} ^ {m_t \times K} $. A tracker maintains a set of trajectories and uses the detections at each timestep to update them. After processing the whole sequence, the system obtains $N_{obj}$ trajectories $\{ T_{j}\in\mathbb{R}^{L_{j}\times K}, j \in N_{obj} \}$ with unique object identities (IDs) where $L_j$ is the track length. For more details, please refer to our supplementary material.

\subsection{Tracking-by-detection with a modified BEVFusion}
\label{subsec:baseline}
To build a strong baseline, we adapt BEVFusion~\cite{liu2023bevfusion}, a SOTA multi-modal 3D object detector to our collected dataset and utilize the popular center-based forward tracking~\cite{yin2021center}. An overview of our system is shown in Fig.~\ref{fig:framework}. The detector consists of backbone modules to extract camera and LiDAR features, a view transform module to construct Bird's Eye View (BEV) features from camera features, a convolutional neural network (CNN) to fuse the multi-modal features, and a head module for 3D object detection and attribute prediction. The prediction head in the detector regresses for each object proposal $p$ its 3D location $(x_p, y_p, z_p)$, 3D size $(l_p, h_p, w_p)$, yaw angle $\theta_p$, and 3D inter-frame offset $O_p$. Thus $K$ in Sec.~\ref{subsec:overview} is 10. The offset prediction is similar to the velocity prediction in~\cite{liu2023bevfusion}, yet we predict the object displacements rather than velocities to utilize training data that have varying frame rates. The ground truth offset is defined as a vector from an object's current location to that in the previous input frame as $O_p = (x_p^{t-1} - x_p^{t}, y_p^{t-1} - y_p^{t}, z_p^{t-1} - z_p^{t})$ . In inference, the predicted offsets are used as learned inter-frame associations to match the detected objects to the trajectories using a greedy assignment strategy~\cite{yin2021center}.

In experiments, we found that the original Lift-Splat-Shoot (LSS) module~\cite{philion2020lift} used in~\cite{liu2023bevfusion} can usually predict unreliable depth, as was also reported in~\cite{li2023bevdepth}. Such depth prediction errors are even more severe for small objects like pedestrians and cause a misalignment of camera and LiDAR features in the 3D space. To avoid this problem, we replace the LSS module with a backward view-transform module. This module builds dense voxels in the 3D space and retrieves camera features by re-projecting the voxels to the images and sampling the 2D features as the voxel features.   

\subsection{Density-aware weighting for tackling the dataset bias}
During data collection, we found that the density of pedestrians in urban driving scenes follows a long-tailed distribution. Crowded scenes are less common compared to scenes that have well-separated objects. A deep-learning model can exploit this data bias~\cite{li2019repair}, overfit to the regular scenes, and fail in the challenging cluttered scenes. 

\begin{figure}[h]
\centering
\includegraphics[width=0.45\textwidth]{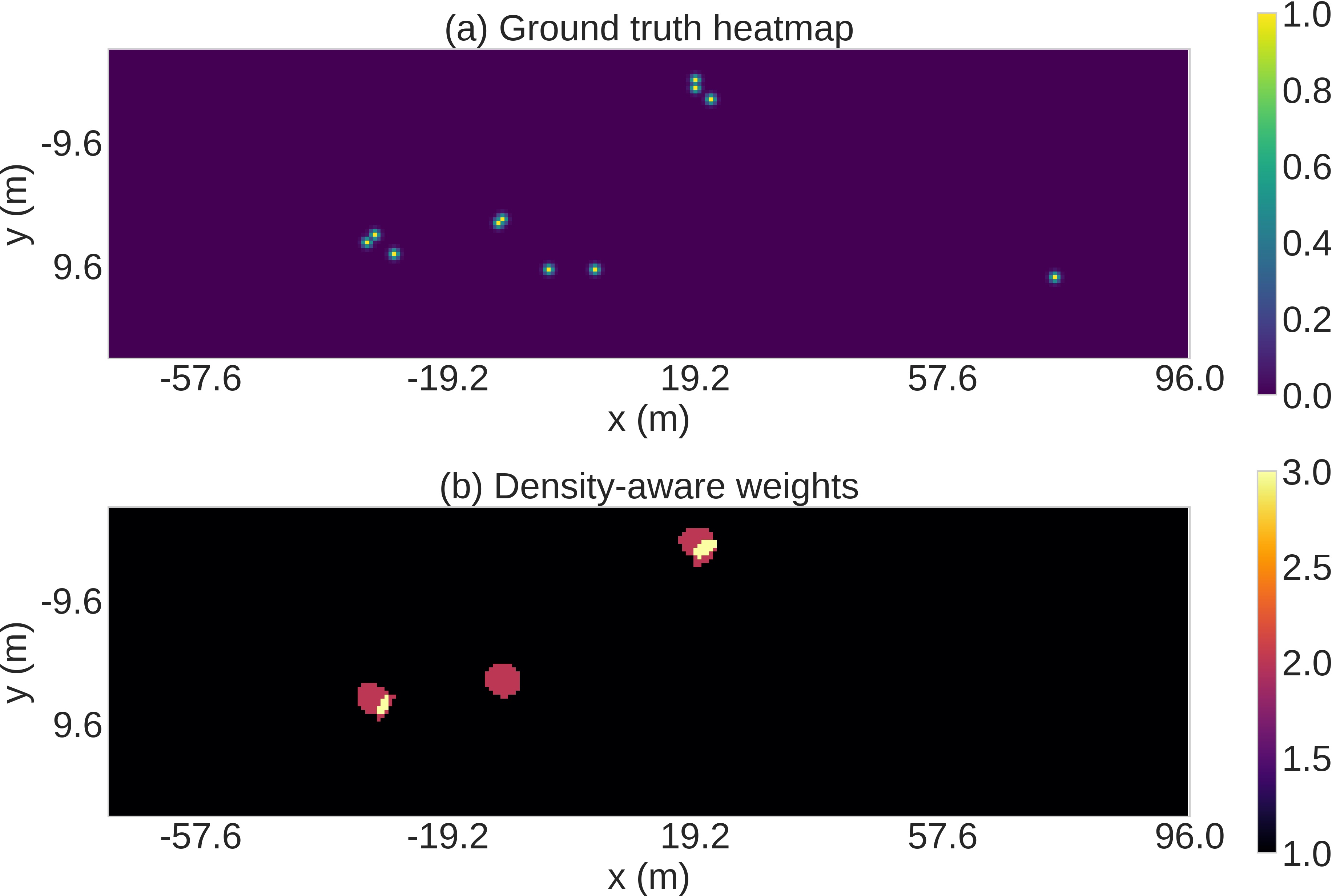}
\caption{(a) A ground truth heatmap and (b) the density-aware weights used for computing the focal loss. The weights are larger in spatial regions where more objects appear.}
\label{fig:density_weights}
\end{figure}

Our baseline uses a location heatmap to supervise the object detection head. Denote the GT location for object $i$ as $(x_i, y_i, z_i)$, and its quantized indices on the BEV heatmap as
\begin{equation}
(j^\star, k^\star) = (\lfloor \frac{x_i - x_{\text{min}}}{\Delta x} \rfloor, \lfloor \frac{y_i - y_{\text{min}}}{\Delta y} \rfloor),    
\end{equation}
where $\Delta x$ and $\Delta y$ are the BEV heatmap resolutions. The GT BEV heatmap is defined as 
\begin{equation}
c_{j,k} = \sum_{i=1}^{N}exp(-\frac{(j - j^\star)^2 + (k - k^\star)^2}{\sigma^2}),
\end{equation} 
where $N$ is the number of GT objects. This GT heatmap is used to penalize the predicted heatmap with a Focal loss~\cite{lin2017focal}. However, we argue that this penalty does not depend on the relative position of the objects. For the input PC of two humans, as long as the predicted heatmap has two activations at the humans' locations, the training detection loss would be the same. In this sense, this supervision does not discriminate a scene of well-separated humans from another scene of closely related humans. To encourage the model to focus on the spatial regions that are more crowded, we propose the density-aware spatial weights (DAW) as 
\begin{equation}
w_{j,k} = \sum_{i=1}^{N}\textbf{1}(\text{dist}(j, k, x_i, y_i) < th),
\end{equation} 
where $\textbf{1}$ is an indicator variable and $\textbf{1}(\text{dist}(j, k, x_i, y_i)) < th$ denotes whether a GT object location is close enough to a heatmap location. The BEV distance $\text{dist}(j, k, x_i, y_i)$ is
\begin{equation}
sqrt((x_{\text{min}} + j\Delta x - x_i)^2 + (y_{\text{min}} + k\Delta y - y_i)^2).
\end{equation} 
The detection loss with DAW at each BEV location is defined as  
\begin{align*}
L_{det} & = \begin{cases}
-w_{j,k}(1 - \hat{p}_{j,k})^\alpha \text{log}(\hat{p}_{j,k}), &c_{j,k}=1,\\
- w_{j,k}(1 - c_{j,k})^\gamma\hat{p}_{j,k}^\alpha \text{log}(1 - \hat{p}_{j,k}), & \text{else},
\end{cases}
\end{align*}
where $\hat{p}_{j,k}$ is the predicted heatmap probability. $\gamma$=4 and $\alpha$=2 are the default parameters. $L_{det}$ reduces to the original focal loss in the baseline~\cite{liu2023bevfusion} when $w_{j,k} = 1$. DAW increases the importance of BEV locations that have pedestrians within a certain range, and the importance scales linearly with the pedestrian number. Fig.~\ref{fig:density_weights} shows an example GT heatmap and the corresponding spatial weights.

\subsection{Discriminating adjacent instances with few LiDAR points using a relationship-aware training objective}
\begin{figure}[h]
\centering
\includegraphics[width=0.48\textwidth]{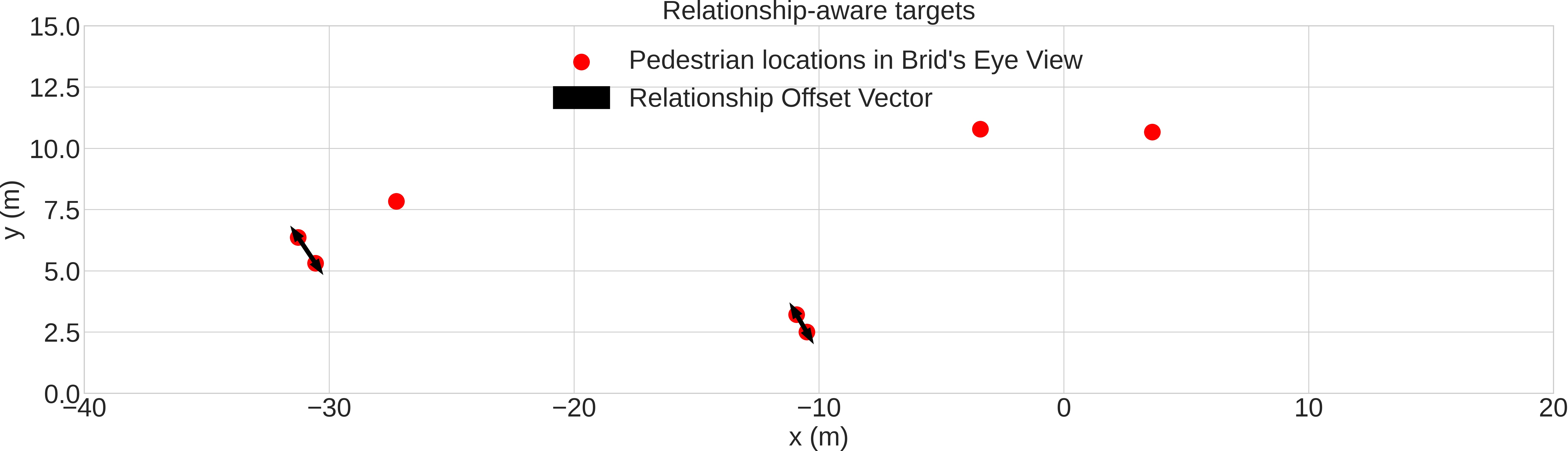}
\caption{Ground truth relationship offset targets shown as arrows. The relationship offset targets differ for adjacent objects, forcing the model to recognize different instances.}
\label{fig:relationship_targets}
\end{figure}

Due to sensor noise and LiDAR PC's sparsity, a pedestrian's PC may only capture a portion of its body. In highly crowded scenes, the PC of different humans are close to each other, and the incomplete PC of two adjacent humans may look like a single human. A model that heavily relies on PC features can fail to distinguish adjacent humans, resulting in erroneous detection and tracking results. To mitigate this problem, we need the detector to learn discriminative visual features to recognize different pedestrians in a crowd. However, the GT heatmaps~\cite{liu2023bevfusion} mentioned above are identical for different humans in a scene, which does not guide the model to recognize the difference between people. To address this gap, we propose a new regression target supervising the detector to learn representations that can discriminate adjacent instances. Specifically, we require the detector to further predict for those pedestrians that have neighbors in 3 meters, a relationship offset vector pointing to its nearest neighbor. The relationship offset vector for a GT pedestrian is,
$
\text{re}_{i} = (x_j - x_i, y_j - y_i),
$
where
\begin{equation}
j = \arg\min_{j \neq i}(x_j - x_i)^2 + (y_j - y_i)^2.
\end{equation} 
This new target increases $K$ in Sec.~\ref{subsec:baseline} from 10 to 12. As shown in Fig.\ref{fig:relationship_targets}, the relationship offset targets differ for two neighboring pedestrians. Thus it penalizes the model for over-fitting to PC features and discarding visual cues to discriminate adjacent pedestrians.

\subsection{Learning High-resolution Sparse Representations}
\begin{figure}[h]
\centering
\includegraphics[width=0.5\textwidth]{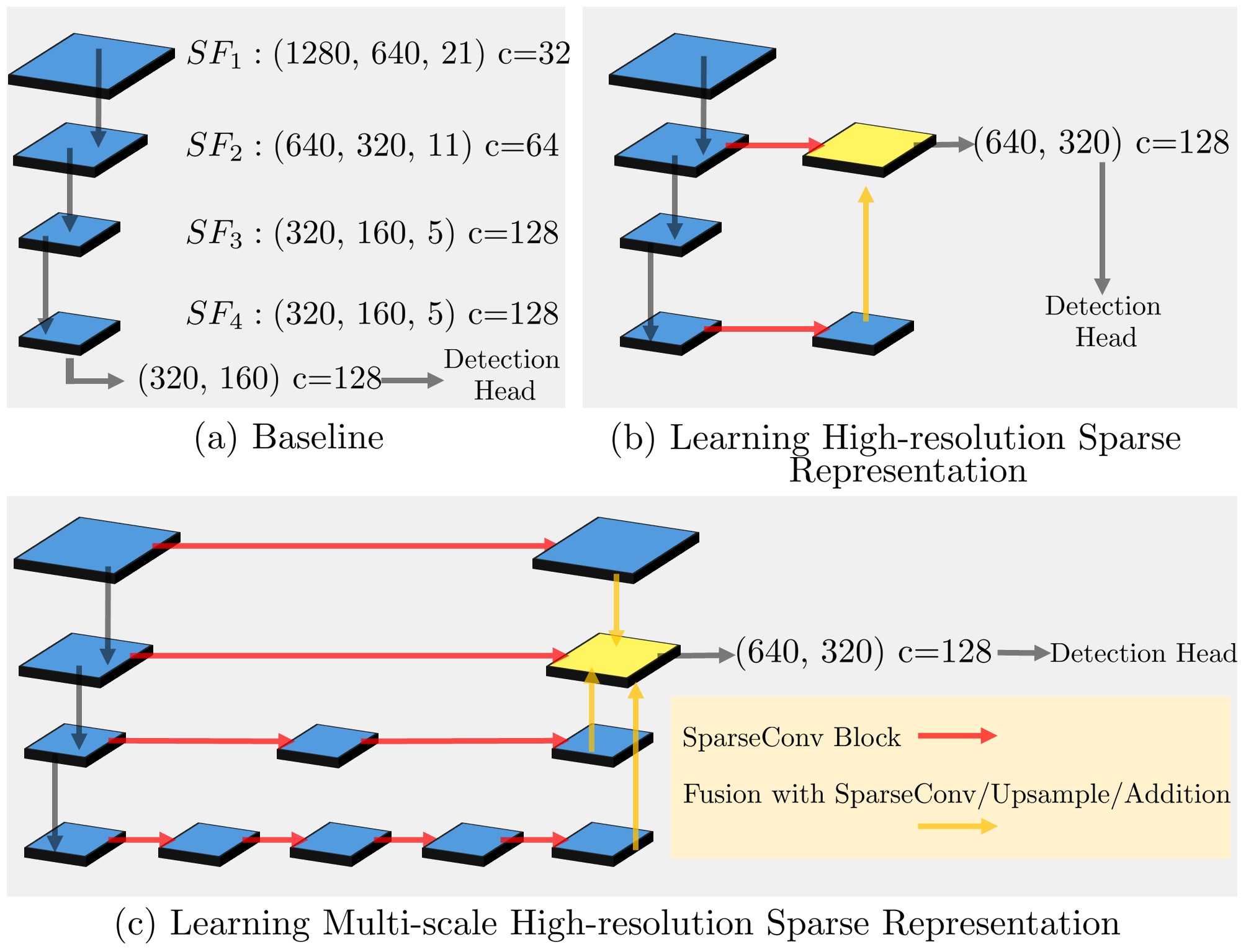}
\caption{We enhance system performance with a high-resolution point cloud representation learning strategy. (a) The baseline point cloud encoder in~\cite{liu2023bevfusion} produces 4 sparse features (SFs) with decreasing spatial shapes shown in the parentheses. (b) We fuse $SF_{2}$ and $SF_{4}$ to produce a high-resolution BEV representation for the detection head. (c) The approach in (b) is further improved to a multi-scale structure.}
\label{fig:hr_representation}
\end{figure}

To further improve the model representation power to detect cluttered small objects, we propose taking advantage of offboard systems' more abundant computational power to learn high-resolution (HR) PC features. The baseline sparse feature encoder is shown in Fig.~\ref{fig:hr_representation} (a), which produces PC features of four resolutions, and the last high-level feature is used for detection. For HR representation learning, we designed two versions that add different computational costs. As shown in Fig.~\ref{fig:hr_representation} (b), instead of utilizing only the low-resolution (LR) features for detection, we fuse the HR PC feature with the LR PC features with several sparse convolutional layers~\cite{yan2018second}. This version adds minor costs and increases the output spatial resolution ($\Delta x$, $\Delta y$) from 0.6m to 0.3m. The more costly multi-scale (MS) version in Fig.~\ref{fig:hr_representation} (c) is inspired by~\cite{wang2020deep}, but our building block is sparse convolutional layers rather than the dense ones in~\cite{wang2020deep} used for visual recognition. It exploits even higher-resolution sparse features and uses more cross-resolution fusion modules to achieve a stronger combination of low-level and high-level sparse features. 

\section{Experiments}
\subsection{Benchmarks and metrics}
For an in-depth performance analysis of offboard 3D MOT of highly crowded pedestrians, we train the detector using our annotated training split and evaluate on PCP-MV. To demonstrate the generality of our system design, we also perform training on nuScenes and follow Chiu et al.~\cite{chiu2021probabilistic} to evaluate on the nuScenes validation set.

Following the pioneer study~\cite{linder2016multi}, we adopt the commonly-used \textit{Multi-Object Tracking Accuracy} (MOTA) to measure the system performance on our PCP-MV. Specifically, it is defined in~\cite{milan2016mot16} as:
\begin{equation}
    \resizebox{.5\hsize}{!}{$\mathbf{MOTA} = 1 -  \frac{IDS + FP + FN}{P},$}
\end{equation}
where $IDS$, $FP$, $FN$, and $P$ are the numbers of identity switches, false positives, false negatives, and GTs, respectively. The evaluation toolkit is adapted from the open-source tool\footnote{https://github.com/JonathonLuiten/TrackEval
}. During matching the predictions and ground truths, the distance metric is BEV intersection over union (IoU), and a threshold of 0.5 is used. For evaluation on the nuScenes validation set, we use the official \textit{Average Multi-Object Tracking Accuracy} (AMOTA) metric and the official evaluation tool.

\begin{table*}[h]
\begin{center}
\begin{tabular}{llll|l|l|l|l|l}
\hline
\multicolumn{4}{c|}{Experimental Setting} &
  \multirow{2}{*}{Extra Computation} &
  \multirow{2}{*}{MOTA$\uparrow$} &
  \multirow{2}{*}{MTR$\uparrow$} &
  \multirow{2}{*}{MLR$\downarrow$} &
  \multirow{2}{*}{IDS$\downarrow$} \\ \cline{1-4}
\multicolumn{1}{l|}{Name} &
  \multicolumn{1}{c|}{DAW} &
  \multicolumn{1}{c|}{RAT} &
  \multicolumn{1}{c|}{HRL} &
   &
   &
   &
   &
   \\ \hline
\multicolumn{1}{l|}{Baseline~\cite{liu2023bevfusion}} &
  \multicolumn{1}{l|}{\xmark} &
  \multicolumn{1}{l|}{\xmark} &
  \xmark &
  \xmark &
  0.172 &
  0.264 &
  0.307 &
  2945 \\
\multicolumn{1}{l|}{+DW} &
  \multicolumn{1}{l|}{\cmark} &
  \multicolumn{1}{l|}{\xmark} &
  \xmark &
  \xmark &
  \te{0.197}{14.5} &
  0.272 &
  0.307 &
  2627 \\
\multicolumn{1}{l|}{+RT} &
  \multicolumn{1}{l|}{\xmark} &
  \multicolumn{1}{l|}{\cmark} &
  \xmark &
  \xmark &
  \te{0.185}{7.5} &
  0.279 &
  0.291 &
  2877 \\
\multicolumn{1}{l|}{+HR} &
  \multicolumn{1}{l|}{\xmark} &
  \multicolumn{1}{l|}{\xmark} &
  \cmark &
  \cmark &
  \te{0.325}{88.9} &
  0.342 &
  0.239 &
  2657 \\
\multicolumn{1}{l|}{+HR+DW} &
  \multicolumn{1}{l|}{\cmark} &
  \multicolumn{1}{l|}{\xmark} &
  \cmark &
  \cmark &
  \te{0.330}{91.8} &
  0.354 &
  0.248 &
  2582 \\
\multicolumn{1}{l|}{Ours (Full)} &
  \multicolumn{1}{l|}{\cmark} &
  \multicolumn{1}{l|}{\cmark} &
  \cmark &
  \cmark &
  \te{0.333}{93.6} &
  0.371 &
  0.229 &
  2460 \\
\multicolumn{1}{l|}{Ours (Full-MS)} &
  \multicolumn{1}{l|}{\cmark} &
  \multicolumn{1}{l|}{\cmark} &
  \cmark &
  \cmark &
  \te{\textbf{0.353}}{105.2} &
  0.315 &
  0.268 &
  2774 \\ \hline
\end{tabular}
\end{center}
\caption{Multi-Object Tracking Accuracy (MOTA) on PCP-MV for pedestrians. Other accompanying metrics in~\cite{milan2016mot16} are also reported. In each experiment, \cmark~indicates a proposed technique is used and \xmark~indicates otherwise. DAW: Density-aware weighting. RAT: Relationship-aware offset targets. HRL: High-resolution representation learning.}
\label{tab:ablation_study}
\end{table*}

\begin{table*}
\footnotesize
\begin{center}
\begin{tabular}{ l|l|cccccccc}
  \hline
  Tracking method & Modalities &  \textbf{Overall} & bicycle & bus & car & motorcycle & \textbf{pedestrian} & trailer & truck \\
  \hline
  \hline
  AB3DMOT~\cite{weng20203d} & LiDAR &  17.9 & 0.9 & 48.9 & 36.0 & 5.1 & 9.1 & 11.1 & 14.2 \\
  \hline
  CenterPoint~\cite{yin2021center} & LiDAR & 65.9	& 43.7 & 80.2 & 84.2 & 59.2 & 77.3 & 51.5 & 65.4 \\
  \hline
  Chiu et al.~\cite{chiu2021probabilistic} & LiDAR + Camera & 68.7 & 49.0 & 82.0 & 84.3 & 70.2 & 76.6 & 53.4 & 65.4 \\
  \hline
  VoxelNeXt~\cite{chen2023voxelnext} & LiDAR & 70.2 & - & - & - & - & - & - & - \\
  \hline  
  Ours & LiDAR + Camera &  \textbf{72.2} & \textbf{64.2} & \textbf{83.0} & \textbf{85.6} & \textbf{75.6} & \textbf{77.9} & 49.6 & \textbf{69.7} \\
  \hline
\end{tabular}
\caption{Overall AMOTA and individual AMOTA for each object category on the nuScenes~\cite{caesar2020nuscenes} validation set.}
\label{tab:quantitative_comp}
\end{center}
\end{table*}

\subsection{Implementation details}
 The model architecture and hyper-parameters inherit the official open-source repository of BEVFusion~\cite{liu2023bevfusion}. Swin Transformer~\cite{liu2021swin} and SECOND~\cite{yan2018second} backbone are used for camera and LiDAR feature extraction respectively. 
 
 For evaluation on PCP-MV, the detection range of our system is $x \in [-96\text{m}, 96\text{m}]$, $y \in [-48\text{m}, 48\text{m}]$, and $z \in [-5\text{m}, 3\text{m}]$. The input PC concatenates data from two frames. The PC is augmented with a time channel representing the current frame (value 0) and the previous frame (value 1). The voxel size for voxelizing the input point cloud is $[\delta x, \delta y, \delta z] = [\text{7.5cm}, \text{7.5cm}, \text{20cm}]$. The voxel size for building the camera features is 60cm for each direction, and a max-pooling operator is used to aggregate MV camera features. In all experiments, we train our model with an AdamW optimizer for 30 epochs with a weight decay of 0.01. The initial learning rate is 0.001 and is multiplied by 0.8 after every 5 epochs. The model is trained on 16 NVIDIA A100 GPUs with batch size 16, and the training takes 12 hours. For training on nuScenes we re-use the hyper-parameters of \cite{liu2023bevfusion}.

\begin{figure*}
\centering
\includegraphics[width=1.\textwidth]{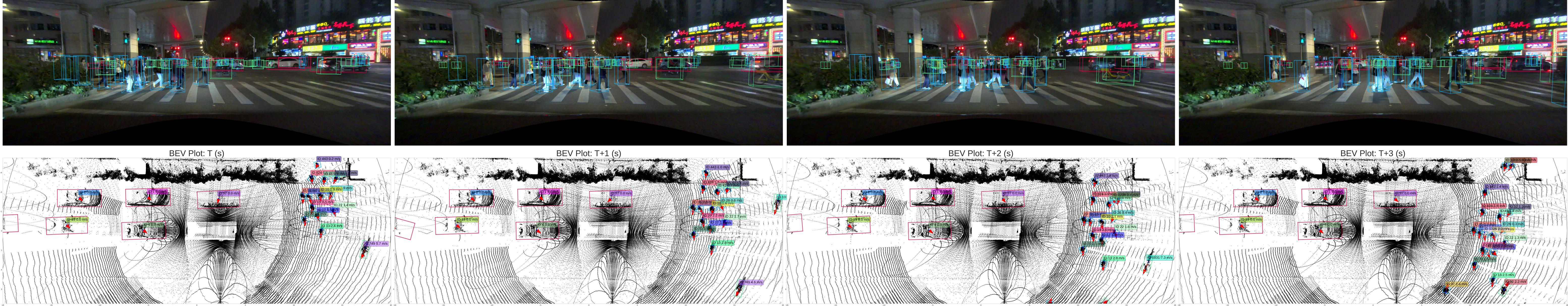}
\caption{Qualitative results of 3D multiple-pedestrian tracking in crowded environments. Detected objects in the video sequence are shown as colored bounding boxes in the BEV plots. Predictions for the pedestrians are shown in blue. Each tracked object has a string indicating its velocity prediction and identity. Better viewed in color and zoom in for more details.}
\label{fig:qualitative}
\end{figure*}

\subsection{Quantitative results and ablation studies}
To validate the effectiveness of our proposed methodology, we conduct extensive ablation studies to demonstrate the contribution of each proposed technique, whose results are summarized in Tab.~\ref{tab:ablation_study}. In these studies, all other experimental configurations are kept the same. The baseline system without our proposed methods achieves a MOTA of 0.172. Despite our baseline taking advantage of recent approaches~\cite{liu2023bevfusion} that perform well on nuScenes, this unsatisfactory performance indicates that PCP-MV is very challenging. 

\noindent\textbf{Effectiveness of the density-aware weighting strategy.} DAW brings a relative improvement of MOTA by 14.5\%. This improvement verifies our hypothesis of the dataset bias. By guiding the model to focus on high-density regions during training, it generalizes better to unseen crowded scenes.

\begin{figure}[h]
\centering
\includegraphics[width=0.48\textwidth]{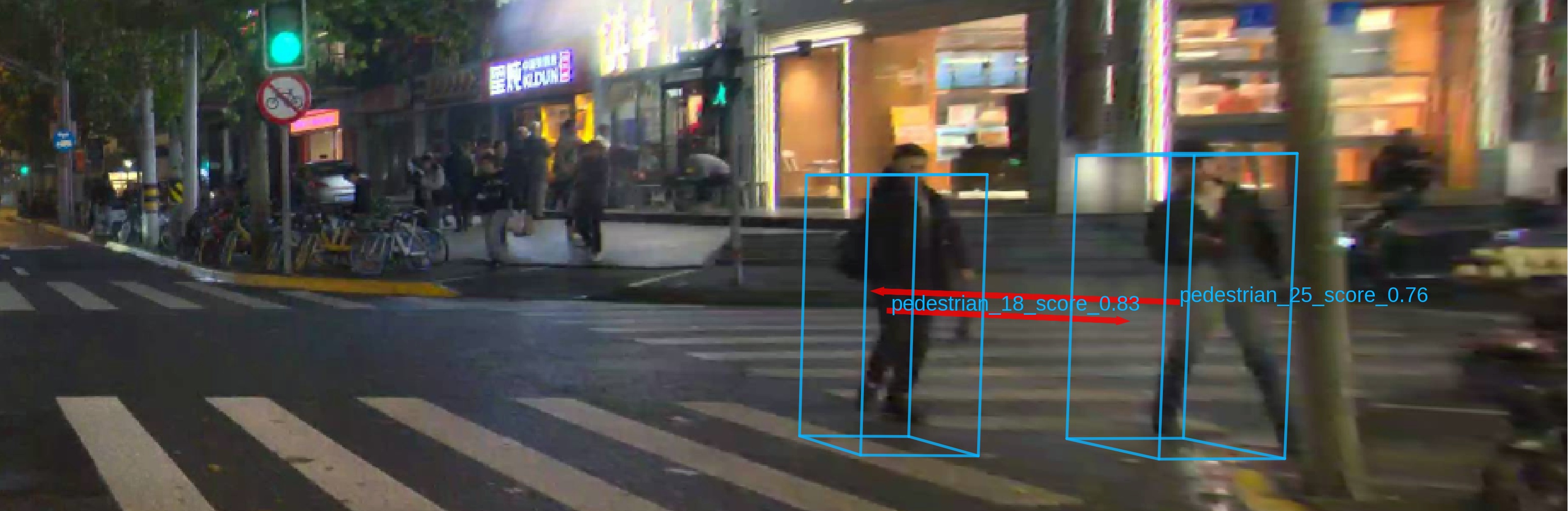}
\caption{Predicted relationship vectors on PCP-MV. An arrow indicates the predicted nearest neighbor for a pedestrian.}
\label{fig:relationship_predictions}
\end{figure}

\noindent\textbf{Learning relationship-aware representations is effective.} Adding the relationship-aware targets improves the MOTA moderately by 7.5\%. The improvement shows that encouraging the model to discriminate adjacent pedestrians can benefit the overall performance. Fig.~\ref{fig:relationship_predictions} shows an example of the predicted relationships on PCP-MV, where the arrows indicate the predicted nearest neighbor for each pedestrian. 

\noindent\textbf{Utilizing high-resolution sparse features is helpful.} The former two strategies can be applied only during training, and do not incur any increased inference cost. When more abundant computational power is available, introducing HR representation learning can bring a leap of MOTA from 0.172 to 0.325. This significant performance boost shows that learning HR features is critical for detecting and tracking small objects such as pedestrians.  

\noindent\textbf{Using multi-scale sparse features is effective.} Compared to our low-cost version (Full) in Fig.~\ref{fig:hr_representation} (b), our multi-scale version (Full-MS) in Fig.~\ref{fig:hr_representation} (c) enjoys a further improvement of MOTA thanks to the improved network design.

\noindent\textbf{The improvements are orthogonal.} When DAW and RAT are gradually added to the system with only HRL, the performance can be further improved. These facts indicate that the improvements brought by learning density-aware and relationship-aware representations are orthogonal to those brought by learning a higher-resolution representation. The two training strategies are still effective for the high-resolution model. 

The results in our final system are significantly improved compared to the baseline. Fig.~\ref{fig:qualitative} shows a qualitative example of the 3D object detection and tracking results of our final system (Full-MS). In this scene with crowded moving pedestrians, our system reliably reconstructs the 3D trajectories of pedestrians with temporally coherent IDs. Such results are used in our cloud-based labeling service and greatly reduce the time spent on manual labeling.

\subsection{Quantitative comparison in less crowded scenes}
While this study focuses on the challenging problem of highly crowded pedestrians, the system also applies to the less crowded scenes. Tab.~\ref{tab:quantitative_comp} compares the AMOTA on the nuScenes validation set with several previous approaches. In terms of overall AMOTA and AMOTA for the pedestrian category, our system achieves higher performance than the classical LiDAR-only approach CenterPoint~\cite{yin2021center}, as well as the LiDAR-camera probabilistic tracking approach of Chiu et al~\cite{chiu2021probabilistic}. Compared with the recent approach VoxelNext~\cite{chen2023voxelnext}, our system also achieves better overall performance. These results indicate our system's competence compared to these studies, validating the generality of our design.

\section{Conclusion}
We conduct an in-depth study of the problem of 3D tracking of pedestrians in high-crowded urban driving scenes. We collect a suitable benchmark and build a strong tracking-by-detection baseline with recent approaches. We achieve significant performance improvements by proposing new model supervision approaches and a high-resolution representation learning strategy. Future studies may consider semi-supervised learning to utilize unlabeled data during training.

\clearpage
\bibliographystyle{IEEEtran}
\bibliography{IEEEabrv, reference}

\end{document}